\documentclass[conference]{IEEEtran}
\IEEEoverridecommandlockouts
\usepackage{cite}
\usepackage{amsmath,amssymb,amsfonts}
\usepackage{algorithmic}
\usepackage{graphicx}
\usepackage{textcomp}
\usepackage{bm}
\usepackage{tikz}
\usepackage{svg}
\usetikzlibrary{arrows.meta, positioning}
\def\BibTeX{{\rm B\kern-.05em{\sc i\kern-.025em b}\kern-.08em
    T\kern-.1667em\lower.7ex\hbox{E}\kern-.125emX}}
\makeatletter
\newcommand{\linebreakand}{%
  \end{@IEEEauthorhalign}
  \hfill\mbox{}\par
  \mbox{}\hfill\begin{@IEEEauthorhalign}
}
\makeatother

\usepackage{color}
\definecolor{DarkBlue}{RGB}{0,0,139}
\definecolor{Green}{RGB}{0,128,0}
\definecolor{FireBrick}{RGB}{178,34,34}
\definecolor{RO}{RGB}{178,34,34}
\definecolor{Black}{RGB}{0,0,0}
\definecolor{Magenta}{RGB}{236,0,141}
\definecolor{Orange}{RGB}{255,128,0}
\definecolor{White}{RGB}{255,255,255}

\usepackage{hyperref} 
\hypersetup{pdfpagemode={UseOutlines},
bookmarksopen=true,
bookmarksopenlevel=0,
hypertexnames=false,
allcolors=DarkBlue, 
colorlinks=true,
pdfstartview={FitV},
unicode,
breaklinks=true,
}
\usepackage{epstopdf}
\graphicspath{{Figures_XX-24/}}
\usepackage{url,comment,balance,bm}
\usepackage{tabularx,adjustbox,booktabs,dcolumn,multirow,array,xfrac,threeparttable}
\newcolumntype{C}[1]{>{\centering\arraybackslash}m{#1}}
\newcolumntype{R}[1]{>{\raggedleft\arraybackslash}m{#1}}
\newcolumntype{L}[1]{>{\raggedright\arraybackslash}m{#1}}
\newcolumntype{d}[1]{D{.}{\cdot}{#1} }
\newcolumntype{$}{>{\global\let\currentrowstyle\relax}}
\newcolumntype{^}{>{\currentrowstyle}}
\newcommand{\rowstyle}[1]{\gdef\currentrowstyle{#1}#1\ignorespaces}
\usepackage[caption=false,font=footnotesize]{subfig} 
\usepackage{gensymb,euscript,units,mathrsfs,bbm,latexsym}

\usepackage[normalem]{ulem}
\usepackage[USenglish]{babel}
\usepackage[calc,useregional]{datetime2}
\DTMlangsetup{monthyearsep={,\space}}
\usepackage[utf8]{inputenc} 
\usepackage[T1]{fontenc}
\usepackage{float}\restylefloat{figure}
\makeatletter
\newcommand*{\@rowstyle}{}
\newcolumntype{=}{
>{\gdef\@rowstyle{}}%
}
\newcolumntype{+}{
>{\@rowstyle}%
}

\definecolor{RO}{RGB}{0,0,0}

\begin{document}

\title{Advanced Deep Learning Approaches for Automated Recognition of Cuneiform Symbols}

\author{
    \IEEEauthorblockN{1\textsuperscript{st} Shahad Elshehaby}
    \IEEEauthorblockA{\textit{College of Engineering and IT} \\
    \textit{University of Dubai}\\
    Dubai, United Arab Emirates\\
    s0000002884@ud.ac.ae}
    \and
    \IEEEauthorblockN{2\textsuperscript{nd} Alavikunhu Panthakkan}
    \IEEEauthorblockA{\textit{College of Engineering and IT} \\
    \textit{University of Dubai}\\
    Dubai, United Arab Emirates\\
    apanthakkan@ud.ac.ae}
    \linebreakand
    \IEEEauthorblockN{3\textsuperscript{rd} Hussain Al-Ahmad}
    \IEEEauthorblockA{\textit{College of Engineering and IT} \\
    \textit{University of Dubai}\\
    Dubai, United Arab Emirates\\
    halahmad@ud.ac.ae}
    \and
    \IEEEauthorblockN{4\textsuperscript{th} Mina Al-Saad}
    \IEEEauthorblockA{\textit{College of Engineering and IT} \\
    \textit{University of Dubai}\\
    Dubai, United Arab Emirates\\
    malsaad@ud.ac.ae}
}

\maketitle

\begin{abstract}
This paper presents a thoroughly automated method for identifying and interpreting cuneiform characters via advanced deep-learning algorithms. Five distinct deep-learning models were trained on a comprehensive dataset of cuneiform characters and evaluated according to critical performance metrics, including accuracy and precision. Two models demonstrated outstanding performance and were subsequently assessed using cuneiform symbols from the Hammurabi law acquisition, notably Hammurabi Law 1. Each model effectively recognized the relevant Akkadian meanings of the symbols and delivered precise English translations. Future work will investigate ensemble and stacking approaches to optimize performance, utilizing hybrid architectures to improve detection accuracy and reliability. This research explores the linguistic relationships between Akkadian, an ancient Mesopotamian language, and Arabic, emphasizing their historical and cultural linkages. This study demonstrates the capability of deep learning to decipher ancient scripts by merging computational linguistics with archaeology, therefore providing significant insights for the comprehension and conservation of human history.
\end{abstract}


\section{Introduction}
Cuneiform, one of the first documented writing systems, originated in ancient Mesopotamia before 3400 BCE \cite{b1}. Initially utilized by the Sumerians, it was then adopted by the Akkadians, Babylonians, Assyrians, and other civilizations. Cuneiform writing was inscribed onto clay tablets with a reed stylus and progressively evolved to encompass the Sumerian, Akkadian, and Hittite languages. The Hammurabi Code, a significant and renowned cuneiform inscription, originates from circa 1754 BCE. This Babylonian legal document, ascribed to King Hammurabi \cite{b2}, has 282 rules addressing many matters and offers significant insights into the comprehension of legal systems and culture in antiquity.

The Hammurabi Code is written in Akkadian, a language of the Semitic family that is linked to Arabic. The two languages are distantly connected as they both belong to the Afro-Asiatic language family \cite{b3}. It transcends mere linguistic classification, representing a political and cultural concept that originated with the Akkadian civilization and subsequently influenced Near Eastern nations, ultimately leading to the emergence of Arabic civilization. Tracing these linkages facilitates the examination of the emergence of language and civilization within society.

Notwithstanding the historical importance of cuneiform, identifying and interpreting cuneiform symbols continues to be a formidable endeavor due to the intricate nature of the writing and the deterioration of old tablets over time. Cuneiform symbols possess several meanings contingent upon their contextual usage, and the extensive array of signs—many of which remain inadequately comprehended—compounds the challenges in translation. Moreover, the interpretation of cuneiform has historically been a manual and laborious process; there exists an imperative demand for an automated and efficient methodology. The utilization of deep learning to interpret ancient writings has become a groundbreaking field of study. Recent research have shown the efficacy of deep learning models in managing the complex patterns of cuneiform characters. Convolutional neural networks (CNNs) have been utilized for symbol identification and classification applications. A work in \cite{b4} introduced a CNN-based approach for detecting cuneiform signs from annotated 3D representations, utilizing lighting augmentations to enhance detection accuracy. Another study in \cite{b5} investigated the application of CNNs for identifying symbols in fragmented and partial inscriptions, showcasing the model's proficiency in managing damaged artifacts. These developments represent a substantial progression beyond conventional image-processing methods, providing archaeologists and historians with formidable tools for interpreting ancient manuscripts.

In addition to detection, the translation of cuneiform into contemporary languages has also been investigated. The study in \cite{b6} presented a technique for automated transliteration using parallel lines, yielding encouraging outcomes despite constrained data availability. This method constituted a major advancement in reconciling ancient and contemporary languages. In our previous work \cite{b7}, the VGG16 deep learning model was utilized to recognize cuneiform symbols from digitized archeological writings. Transfer learning approaches were employed to improve the model's capacity to accurately depict historical symbols.

Furthermore, additional research has explored the extensive uses of machine learning and deep learning methodologies for the analysis of cuneiform data. For example, \cite{b8} illustrated the efficacy of deep learning in the large-scale categorization of point clouds from cuneiform tablets, attaining notable enhancements in accuracy. Likewise, the research in \cite{b9} utilized machine learning techniques with unigram features on a balanced dataset of cuneiform characters, yielding strong classification outcomes. Collectively, these investigations underscore the advancing function of artificial intelligence in interpreting intricate ancient manuscripts.

This research study presents a thorough methodology for cuneiform symbol recognition utilizing five distinct models: VGG16, EfficientNet, MobileNet, InceptionResNetv2, and 2D CNN models. These models were trained on a particular dataset of cuneiforms and evaluated on symbols derived from the Hammurabi Code. The main objective is to ascertain the Akkadian meaning of each sign together with its respective English translation. The linguistic relationship between Akkadian and Arabic is further examined, enhancing research in archaeology, computational linguistics, and historical studies. 

\section{Dataset}
This cuneiform dataset has 14,100 representations of various cuneiform characters. The program started by loading and extracting 235 distinct cuneiforms extracted from \cite{b10}, with preprocessing yielding 10 varied representations for each. The preprocessing procedures encompass standardizing symbols, shrinking them, and reducing noise to enhance clarity and uniformity. Subsequently, augmentation techniques were employed, adding five more symbols for each character to enhance the quantity and variety of the dataset. The additional dataset presents the models with varied viewpoints of each sign, hence improving their generalization skills across diverse representations.

\begin{table}[htb]
\caption{Dataset Splitting for Cuneiform Character Recognition}
\label{tab:DatasetSplitting}
\begin{adjustbox}{center}
\setlength{\tabcolsep}{9pt}
\begin{tabular}{c c c c}  
\toprule
\rowstyle{\bfseries} 
\textbf{Counting Parameter} & \multicolumn{3}{c}{\textbf{Dataset Splitting}} \\
\cmidrule(lr){2-4}
& \textbf{Testing} & \textbf{Validating} & \textbf{Training} \\
\midrule
\textbf{Percentage (\%)} & 40\% & 24\% & 36\% \\
\textbf{No. of Characters} & 5,640 & 3,384 & 5,076  \\
\bottomrule
\end{tabular}
\end{adjustbox}
\end{table}

The dataset is allocated according to a 40-24-36 distribution for testing, validation, and training, respectively. Subsequently, 5,640 samples were designated for testing, 3,384 samples for validation, and 5,076 samples for training, as illustrated in Table~\ref{tab:DatasetSplitting}. The models are trained on a suitably varied dataset, ensuring adequate data remains for an impartial evaluation of the models.

\section{Methodology}

This section outlines the approach employed by the proposed algorithm for finding and interpreting cuneiform characters. The methodology is designed to optimize the detection precision of cuneiform letters and ascertain their Akkadian meanings, subsequently translating them into English. The approach has multiple essential parts, incorporating the dataset preparation phase completed in the preceding section. These critical phases include model design and architecture, the training process, and assessment procedures. Each phase is crucial for developing a strong and efficient deep learning model in this research.

\begin{figure}[htbp]
\centerline{\includegraphics[width=90mm]{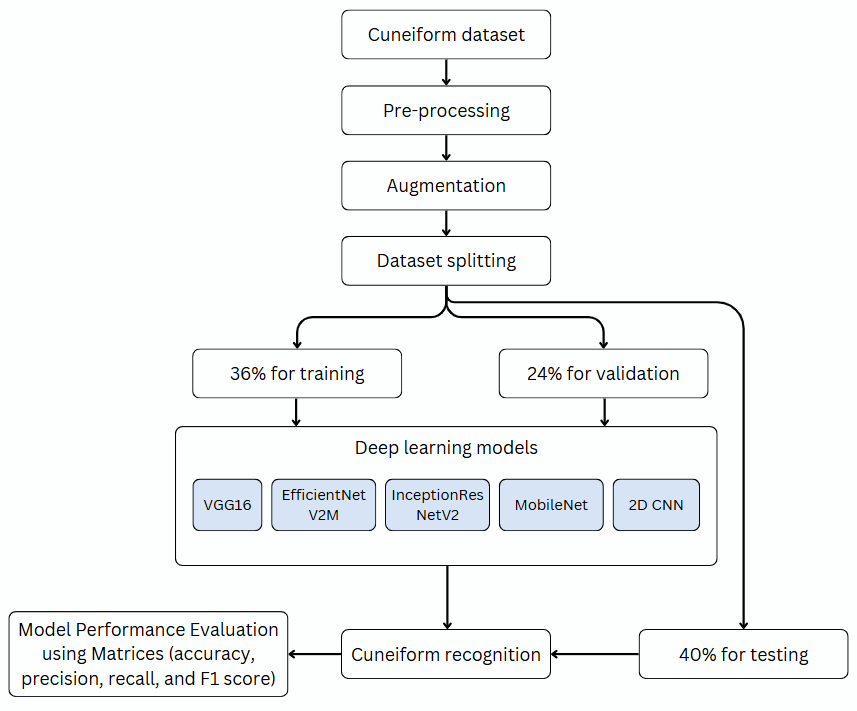}}
\caption{Flowchart of the proposed Cuneiform Recognition methodology}
\label{fig}
\end{figure}

\subsection{Model Design and Architecture}

This study examined five distinct deep learning models, each with distinctive architectures designed to enhance performance in cuneiform detection. The models comprised versions of Convolutional Neural Networks (CNNs), Recurrent Neural Networks (RNNs), and hybrid models that integrate both methodologies \cite{b11}. The models were designed with varying weights and layers, ranging from shallow networks for rapid computing to deeper, more complicated architectures for capturing the subtle properties of cuneiform characters. The architectural decision was influenced by the necessity to balance accuracy with computing performance, given the complexity of the dataset.
\\
\begin{itemize}
\item VGG16 Model: This model utilizes a pre-trained VGG16 architecture, supplemented with extra layers specifically designed for cuneiform symbol identification. The VGG16 model is ideally suited for this task because of its deep architecture, which captures the delicate nuances of cuneiform symbols.

\item EfficientNetV2M: This model is selected for its remarkable equilibrium of precision and efficiency, utilizing a compound scaling approach to appropriately modify depth, breadth, and resolution. This functionality guarantees peak performance while effectively utilizing computing resources, especially for intricate datasets.

\item InceptionResNetV2: This hybrid architecture amalgamates the benefits of Inception and ResNet frameworks, allowing the robust recognition of intricate and varied cuneiform characters. Its multi-scale feature extraction using parallel convolutional pathways improves its capacity to represent complex patterns efficiently.

\item MobileNet: This model was incorporated for its computational efficiency, utilizing depthwise separable convolutions to markedly decrease parameters and calculations. Although it is lightweight, it retains competitive accuracy, rendering it suitable for mobile or resource-limited applications.

\item 2D CNN: A customized 2D CNN was created to particularly capture the spatial and artistic attributes inherent to cuneiform symbols. This design is refined for precise feature extraction, highlighting the unique structural components of the script \cite{b12}. 
\end{itemize}

\subsection{Model Training and Optimization}
Each model experienced training on the selected dataset for 50 epochs, providing enough opportunity to enhance their ability to detect and classify cuneiform characters appropriately. The training approach employed the Adam optimizer \cite{b13}, which integrates an adjustable learning rate with effective management of sparse gradients, a prevalent issue in character recognition applications. To mitigate overfitting, early stopping was employed to assess the model's performance on the validation set and terminate training if no enhancements were seen over five successive epochs (patience = 5). This method mitigates the danger of overfitting while conserving computing resources by eliminating superfluous repeats.

\begin{figure}[htbp]
    \centering
    \includegraphics[width=88mm]{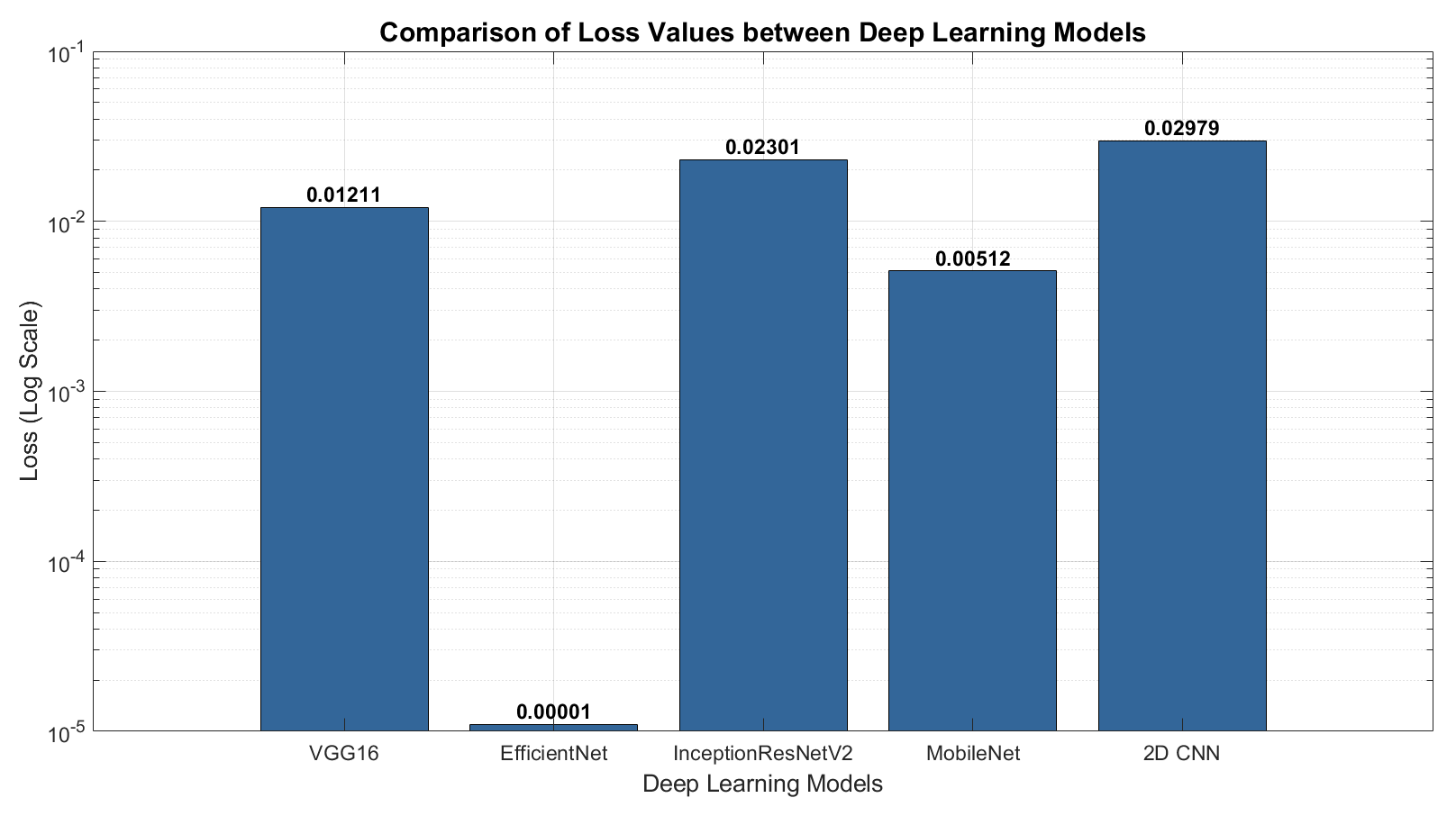}
    \caption{A log-scale comparison of loss values across deep learning models}
    \label{fig:loss}
\end{figure}

Figure~\ref{fig:loss} depicts the loss values for each model during the training process, emphasizing EfficientNet as the superior model with minimal loss, succeeded by VGG16 \cite{b14}. These findings highlight the appropriateness of EfficientNet and VGG16 for more assessment and experimentation, due to their reliable and strong performance.

\subsection{Performance Assessment and Metric Comparison}
The performance of each model was evaluated following training utilizing a supervised learning test set. A thorough assessment of each model's advantages and disadvantages was conducted by calculating essential metrics, including accuracy, precision, recall, and F1 score, as outlined in Equations (1) through (4). These equations delineate the mathematical formulations of the assessment metrics, encompassing the following parameters: True Positive (TP), True Negative (TN), False Positive (FP), and False Negative (FN) \cite{b15}. 

\begin{equation}
\begin{aligned}
Accuracy = \frac{TP+TN}{TP+TN+FP+FN}\label{eq}
\end{aligned}
\end{equation}

\begin{equation}
\begin{aligned}
Precision = \frac{TP}{TP+FP}\label{eq}
\end{aligned}
\end{equation}

\begin{equation}
Recall = \frac{TP}{TP+FN}\label{eq}
\end{equation}

\begin{equation}
F1-score = 2 \times \frac{(Precision \times Recall)}{(Precision + Recall)}\label{eq}
\end{equation}

\begin{table}[htb]
\caption{Deep Learning models performance metrics comparison}
\label{tab:model_performance}
\begin{adjustbox}{center}
\setlength{\tabcolsep}{4pt} 
\renewcommand{\arraystretch}{1.5} 
\begin{tabular}{l|c|c|c|c} 
\toprule
\textbf{Model} & \textbf{Accuracy} & \textbf{Precision} & \textbf{Recall} & \textbf{F1 Score} \\
\midrule
VGG16 & 0.9994 & 0.9993 & 0.9995 & 0.9996 \\
EfficientNet & 0.9999 & 0.9998 & 0.9998 & 0.9999 \\
MobileNet & 0.9991 & 0.9993 & 0.9991 & 0.9993 \\
Inception-\newline ResNetV2 & 0.9857 & 0.9857 & 0.9855 & 0.9855 \\
2D CNN & 0.9875 & 0.9934 & 0.9921 & 0.9914 \\
\bottomrule
\end{tabular}
\end{adjustbox}
\end{table}

\section{Comparative Linguistic Investigation}
In addition to model training and assessment, a linguistic analysis was performed to examine the relationships between the Akkadian and Arabic languages. The linguistic investigation encompassed the examination of identified cuneiform symbols and their Akkadian interpretations, framed within established language similarities with Arabic \cite{b16}. The aim was to ascertain the potential impacts and shared attributes of language that might enhance the validation of the identified translations via a more profound understanding of the historical and cultural connections between these ancient languages \cite{b17}.

\section{Results and Discussion}
This section outlines the results of employing diverse trained models to examine Hammurabi's Law 1 \cite{b17}. The archaic manuscript was supplied as a scanned picture including 35 cuneiform symbols. To enhance model processing, the picture was subjected to a sequence of preprocessing techniques, including scaling, thresholding, dilation, and segmentation to identify distinct characters \cite{b14}. The segmented characteristics were subsequently assessed utilizing five deep learning models: VGG16, EfficientNetV2M, MobileNet, InceptionResNetV2, and a bespoke 2D CNN.

\subsection{Preprocessing and Character Segmentation}
\begin{figure}[htbp]
\centerline{\fbox{\includegraphics[width=85mm]{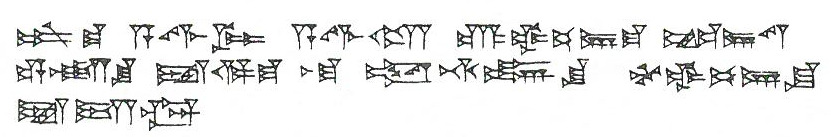}}}
\caption{Scanned image of Hammurabi's Law 1 \cite{b10}}
\label{fig:Law1}
\end{figure}

To provide consistency among models, the picture of Hammurabi's Law 1 depicted in Figure~\ref{fig:Law1} was reduced to a width of 1000 pixels while preserving its aspect ratio. Thresholding transformed the picture into a binary representation, improving character discernibility. Dilation was utilized to enhance clarity, succeeded by contour recognition and sorting for accurate line and character segmentation. Each segmented character was enlarged and re-thresholded to conform to the training data format.

\subsection{Model-Based Character Recognition}
Each character underwent processing via deep learning models to determine its associated cuneiform symbol. The predictions were evaluated against manually chosen ground truth labels for accuracy evaluation. The models' performance was shown and examined to emphasize their prediction potential.

\subsection{Evaluation and Analysis}
\begin{figure}[htbp]
    \centering
    \subfloat[VGG16 model]{%
        \centerline{\fbox{\includegraphics[width=65mm]{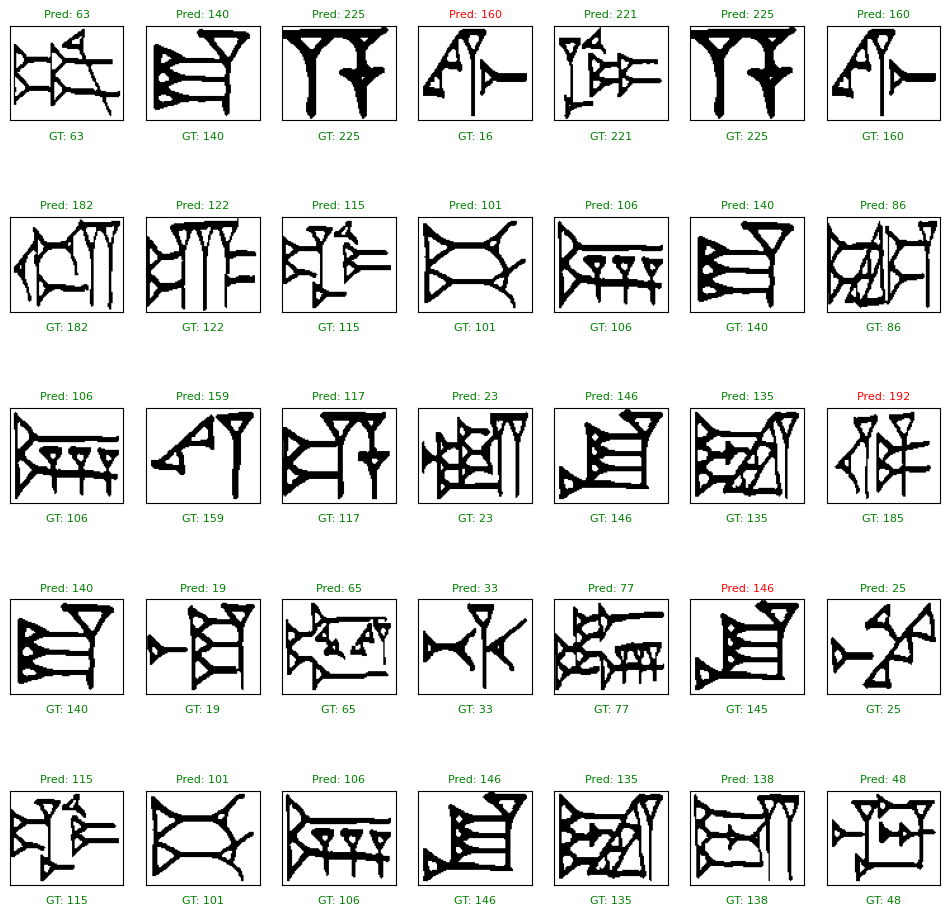}}} 
        \label{fig:site1}
    }
    \hfill
    \subfloat[EfficientNetV2M model]{%
        \centerline{\fbox{\includegraphics[width=65mm]{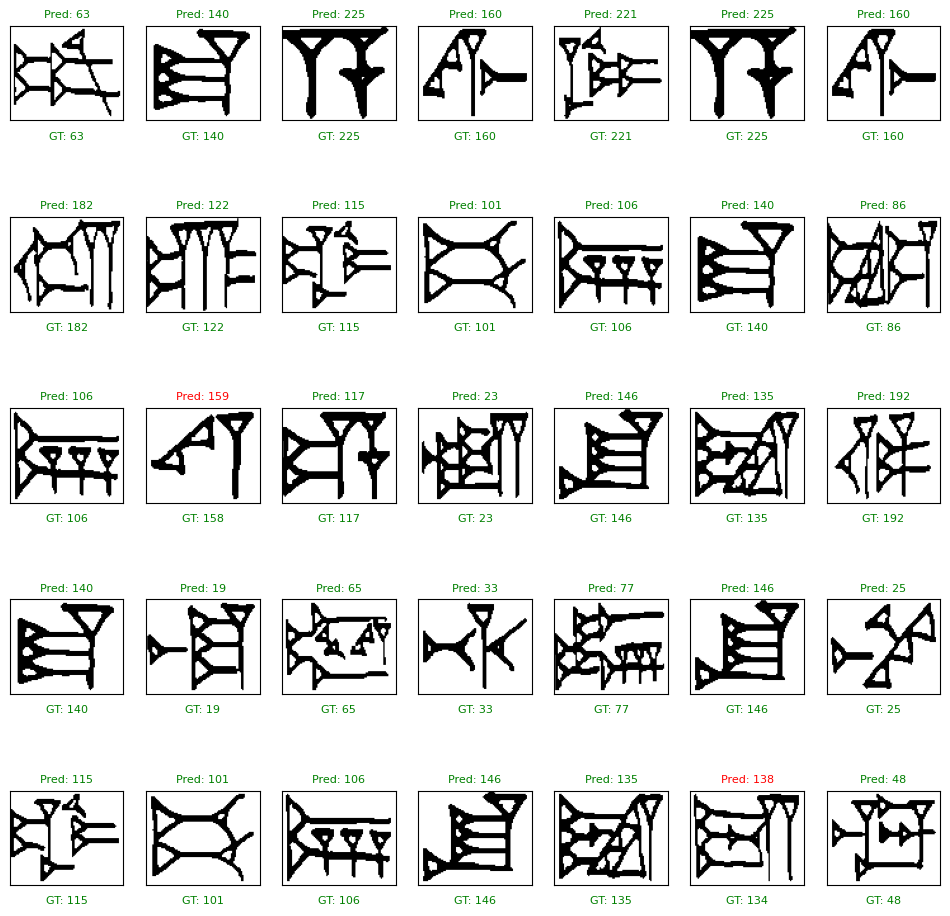}}}
        \label{fig:site3}
    }
    \caption{Prediction results of the best-two performing deep learning models compared with the ground truth data}
    \label{fig:main-figure}
\end{figure}

The outcomes derived from the unsupervised testing juxtapose the anticipated symbols with the actual facts. Multiple models, notably VGG16 and EfficientNetV2M, have demonstrated superior performance, exhibiting a greater accuracy percentage (\%) in predictions. Characters from Hammurabi Law 1 that are accurately identified are highlighted in green. The findings derive from a comparison of the predicted values (displayed on top) with the ground truth values (displayed on the bottom), as seen in Figure~\ref{fig:main-figure}. The VGG16 model had a relative accuracy of 88.87\% when evaluated on the characters from Hammurabi Law 1. Nonetheless, the EfficientNetV2M model had a superior accuracy of 98.31\%. 

\subsection{Cuneiform Character Translation and Linguistic Insights}
\begin{figure}[htbp]
\centerline{\fbox{\includegraphics[width=75mm]{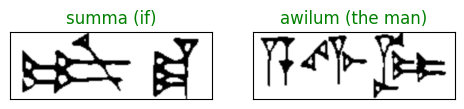}}}
\caption{Sample of the recognized cuneiform words}
\label{fig:words}
\end{figure}

Upon identification, the system correlated the recognized characters with their respective Akkadian and English translations from a predetermined database. The letters were subsequently amalgamated into whole sentences, rebuilding Hammurabi's Law 1. Figure~\ref{fig:words} presents a sample of EfficientNetV2M's predictions, displaying the initial two words of the legislation with their Akkadian and English translations.

A comparative linguistic research revealed correlations between Akkadian and Arabic. Table~\ref{tab:cuneiform_translation} presents cuneiform lexemes with their Arabic transliterations, highlighting parallels in both sound and semantics. These linguistic analogies provide profound insights into the historical history of language and cultural interchange.

\begin{table}[htb]
\caption{Cuneiform words translation sample}
\label{tab:cuneiform_translation}
\begin{adjustbox}{center}
\setlength{\tabcolsep}{4pt} 
\renewcommand{\arraystretch}{1.5} 
\begin{tabular}{l|c|c|c|c} 
\toprule
\textbf{English} & \textbf{Cuneiform} & \textbf{Akkadian} & \textbf{Arabic Transliteration} & \textbf{Arabic} \\
\midrule
if & \includegraphics[width=12mm]{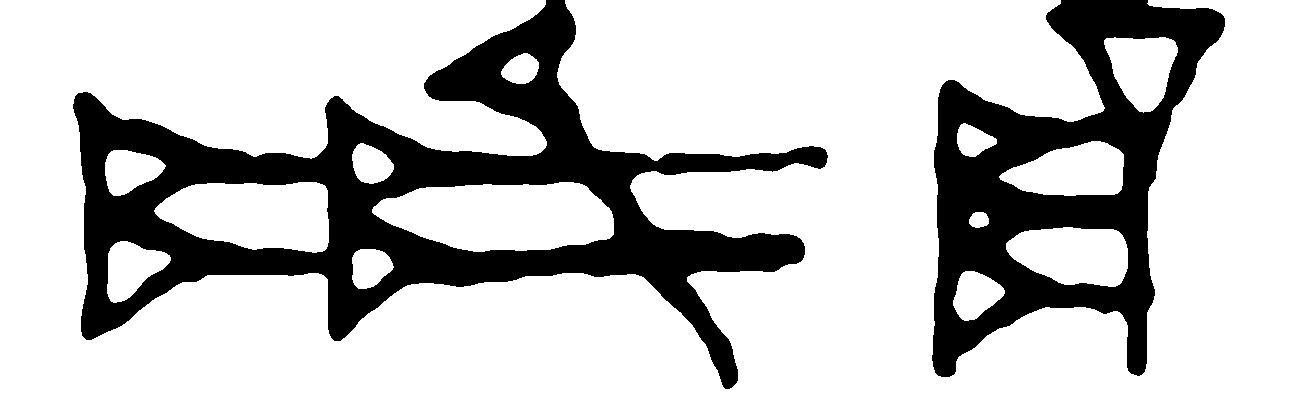} & šumma & \includegraphics[width=4.3mm]{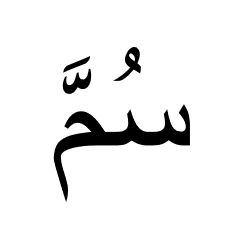} & \includegraphics[width=2.7mm]{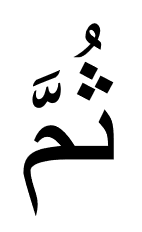} \\
executed & \includegraphics[width=17mm]{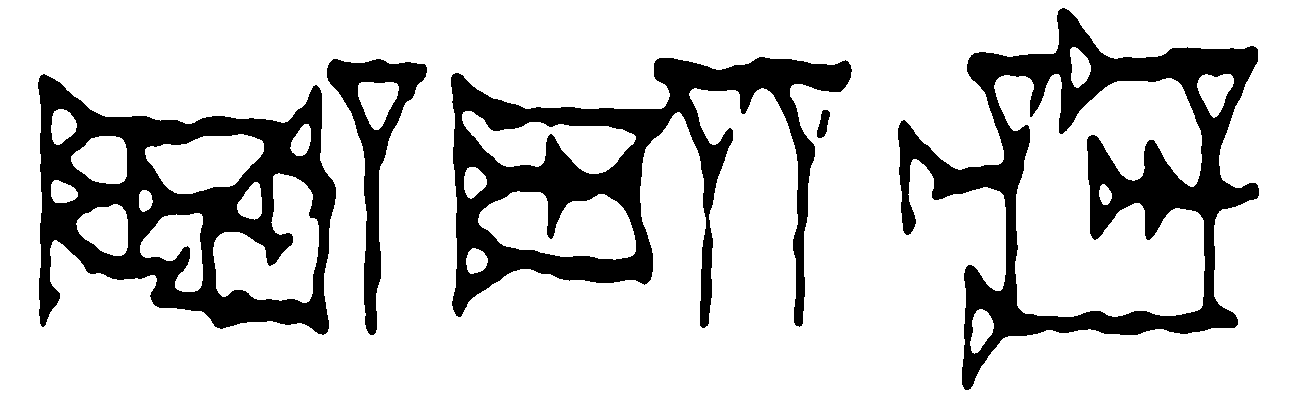} & iddâk & \includegraphics[width=4mm]{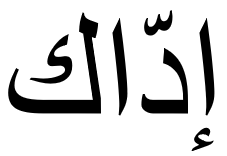} & \includegraphics[width=5mm]{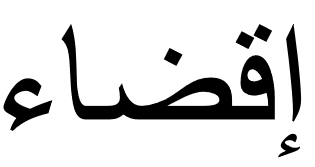} \\
not & \includegraphics[width=12mm]{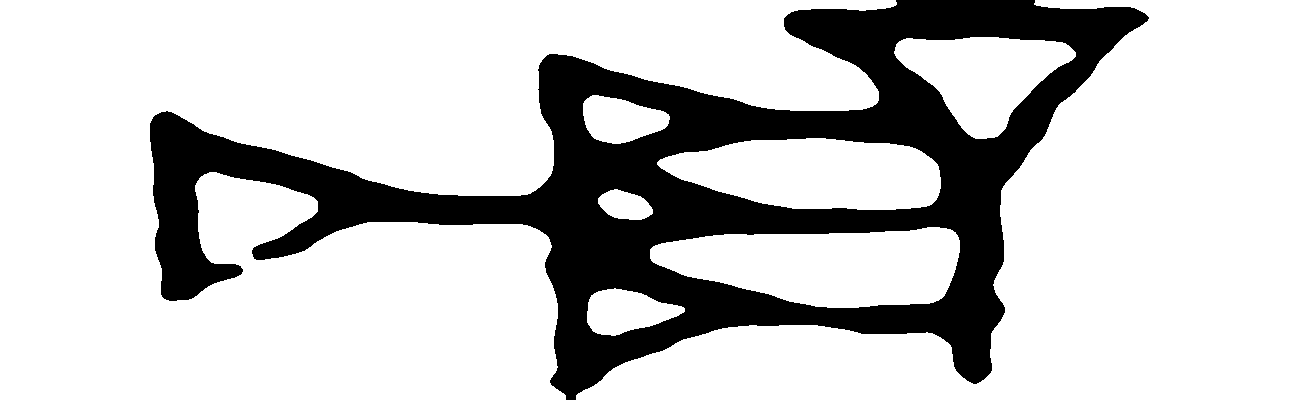} & lā & \includegraphics[width=1.9mm]{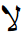} & \includegraphics[width=1.9mm]{Figures/arabic3.png} \\
\bottomrule
\end{tabular}
\end{adjustbox}
\end{table}

\subsection{Observations}
The results of this assessment indicate that the models are proficient in identifying cuneiform characters \cite{b15}, although issues persist in properly predicting specific characters due to their complexity and the similarities across various signs. The identified faults were examined to uncover potential enhancements in preprocessing, model design, or training approach.

The unsupervised evaluation of Hammurabi Law 1 illustrates the practical use of the trained models, revealing that the most favorable outcomes are derived from the two superior models: VGG16 and EfficientNetV2M. Enhanced refinement in the preprocessing pipeline and model training might ultimately improve character identification accuracy, hence benefiting cuneiform translation systems.

\section{Conclusion}
This study has introduced a comprehensive method for the detection and translation of cuneiform characters via deep learning techniques. We chose five distinct models—VGG16, EfficientNet, MobileNet, InceptionResNetV2, and a 2D CNN—trained on a specialized cuneiform dataset. Subsequently, we selected the two best-performing models, which were evaluated on Hammurabi Law 1, and proceeded to identify the Akkadian symbols and their English translations. The research examined the linguistic connections between Akkadian and Arabic, therefore illuminating the historical and cultural evolution of the Afro-Asiatic language family. The findings highlight the efficacy of deep learning in automated cuneiform recognition, advancing the research of ancient writings and diminishing conventional labor-intensive tasks. In the future, models that attain optimal metrics may be chosen for enhancement via a hybrid methodology. Efforts may also focus on the further refining of these models by employing more sophisticated approaches or including other datasets, as well as applying the methodology to other ancient languages, such as Egyptian Hieroglyphs \cite{b18}. This expansion will facilitate extensive historical and archaeological study, so greatly enhancing the comprehension and preservation of human history.
\\

\end{document}